\documentclass[letterpaper, 10 pt, conference]{IEEEtran}  

\IEEEoverridecommandlockouts                              

\usepackage{amsmath}

\makeatletter

\usepackage{amsmath, amssymb, amsfonts}

\usepackage{graphicx} 
\usepackage{pifont}
\usepackage{amsmath}
\usepackage{multirow}
\usepackage{graphicx}
\usepackage{cite}

\title{\LARGE \bf
Multi-Agent Off-World Exploration for Sparse Evidence Discovery via Gaussian Belief Mapping and Dual-Domain Coverage
}

\author{
Zhuoran Qiao$^{1}$,
Tianxin Hu$^{2}$,
Thien-Minh Nguyen$^{3}$,
Shenghai Yuan$^{2}$\thanks{Corresponding author.} \\
\\
$^{1}$National University of Singapore, Singapore \\
$^{2}$Nanyang Technological University, Singapore \\
$^{3}$The University of Queensland, Australia
}

\begin{document}

\maketitle
\thispagestyle{empty}
\pagestyle{empty}

\begin{abstract}
Off-world multi-robot exploration is challenged by sparse targets, limited sensing, hazardous terrain, and restricted communication. Many scientifically valuable clues are visually ambiguous and often require close-range observations, making efficient and safe informative path planning essential. Existing methods often rely on predefined areas of interest (AOIs), which may be incomplete or biased, and typically handle terrain risk only through soft penalties, which are insufficient for avoiding non-recoverable regions. To address these issues, we propose a multi-agent informative path planning framework for sparse evidence discovery based on Gaussian belief mapping and dual-domain coverage. The method maintains Gaussian-process-based interest and risk beliefs and combines them with trajectory-intent representations to support coordinated sequential decision-making among multiple agents. It further prioritizes search inside the AOI while preserving limited exploration outside it, thereby improving robustness to AOI bias. In addition, the risk-aware design helps agents balance information gain and operational safety in hazardous environments. Experimental results in simulated lunar environments show that the proposed method consistently outperforms sampling-based and greedy baselines under different budgets and communication ranges. In particular, it achieves lower final uncertainty in risk-aware settings and remains robust under limited communication, demonstrating its effectiveness for cooperative off-world robotic exploration.
\end{abstract}

\section{INTRODUCTION}
Off-world surface exploration increasingly demands autonomous multi-robot systems that can search for subtle scientific cues under severe mobility hazards and limited sensing. Many high-value targets (e.g., ancient biological relics, biosignature-like cues, or fine-grained geological evidence) are small, visually ambiguous, and often only confirmable at close range. Consequently, an onboard camera typically has a narrow effective sensing footprint, and mission success depends critically on how robots allocate motion to acquire close-up observations rather than on long-range perception alone.

Coverage planning and multi-robot exploration are natural responses to limited sensing footprint: multiple agents can parallelize search and reduce time-to-discovery \cite{viseras/2016/decentralized,corah2017efficient}. However, two practical issues are frequently under-modeled in existing exploration and coverage formulations. First, search is rarely confined to a perfectly defined area of interest (AOI). AOIs are often specified by coarse priors (orbital cues, scientific hypotheses, or operator-defined polygons) and can be incomplete or biased. Optimizing coverage strictly within an AOI can therefore lead to systematic blind spots and reduced robustness when evidence lies outside the presumed region \cite{firstIPP/2013/sampling,hitz/2017/adaptive}. Second, off-world terrains contain non-recoverable hazards and high-slip regions that can trap a rover (entering a region is feasible, but exiting is not reliably achievable). In such settings, soft risk penalties alone are often insufficient: planning must explicitly enforce recoverability to prevent trajectories that can lead to mission-ending states \cite{RIG/2014sampling,MAIPPweakness/2017/randomized}.

\begin{figure}[t]
\centering
\includegraphics[width=1\linewidth, trim=0 70 0 90, clip]{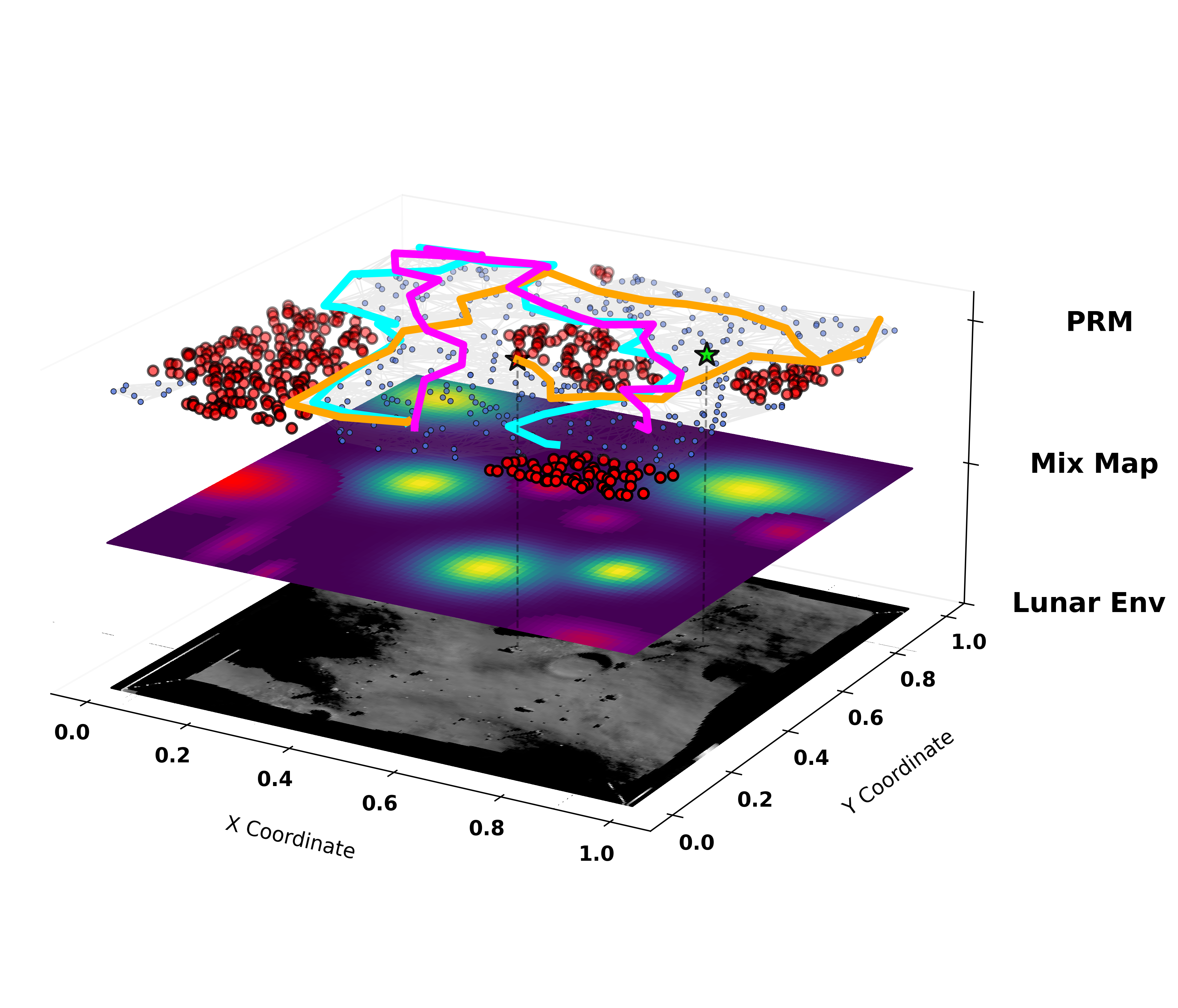}
\vspace{-15pt}
\caption{\textbf{Validation in Lunar Environment 1.} To efficiently explore the AOI in hazardous lunar terrain, three agents are deployed to operate collaboratively in parallel, and their trajectories are shown by the yellow, blue, and purple curves. From bottom to top, the figure presents the original lunar environment, the mixed Gaussian map, and the PRM planning layer. The red topological nodes and Gaussian-shaped distributions represent forbidden regions by imposing a higher traversal cost during path search.} 
\label{fig:motivation}
\vspace{-15pt}
\end{figure}

This paper addresses these gaps in a unified multi-agent visual search framework implemented in a Gazebo off-world simulation environment. We consider a team of robots equipped with onboard cameras that must detect sparse evidence online. Detections are intermittent and spatially uncertain, and thus must be integrated into a representation that is both lightweight and planner-compatible. We maintain a sparse Gaussian evidence belief in the world frame, updated incrementally from onboard visual observations \cite{hitz/2017/adaptive}. This belief supports principled replanning by quantifying where evidence is likely and where uncertainty remains high.

On top of the belief map, we adopt an intent-based multi-agent planning architecture \cite{Intent/MULTI/RL/YANG/2023,deolasee2024dypnipp}. At each replanning cycle, each agent proposes a small set of candidate intents (e.g., evidence chasing, frontier coverage), and a coordinator selects a non-conflicting subset that maximizes team-level marginal utility under motion feasibility and collision avoidance. Crucially, we extend conventional AOI-centric coverage to a dual-domain objective that explicitly allocates search effort both inside and outside the AOI. The AOI is treated as a high-priority domain, while a controlled background coverage budget mitigates prior bias and enables discovery beyond the assumed region.

To ensure operational safety, we incorporate terrain risk and recoverability constraints through a two-stage mechanism: (i) a terrain-derived risk field that discourages hazardous proximity and high-slip regions, and (ii) a hard safety layer that rejects candidate trajectories violating a recoverability criterion defined by dynamic safety buffers and local feasibility checks. This combination prevents "enter-but-not-exit" behaviors that can otherwise arise when the planner trades safety for short-term coverage gain \cite{RIG2/2019sampling,RIG3/2018deterministic}.

We evaluate the proposed system in diverse off-world simulation scenarios with varying hazard layouts, AOI bias levels, and evidence sparsity. Results show that dual-domain exploration improves out-of-AOI discovery and reduces failure modes under AOI misspecification, while recoverability constraints significantly reduce mission-ending traps with minimal loss in search efficiency.
The contribution of this paper can be summarized as:
\begin{itemize}
  \item A multi-agent off-world visual search framework that fuses intermittent detections into a sparse GP-based evidence belief for online replanning.
  \item A dual-domain intent-aware cooperative planning strategy that optimizes coverage inside the AOI and in the background region and leverages trajectory intent to reduce redundant exploration and achieve lower final uncertainty under shared budgets.
  \item A risk-aware belief and decision-making mechanism that maintains a GP-based terrain risk belief and integrates it into planning to improve exploration quality and stability in hazardous environments.
\end{itemize}

\section{Related Works}

\subsection{Adaptive Single-Agent IPP}
Informative Path Planning (IPP) has a long research history. Hollinger and Sukhatme \cite{firstIPP/2013/sampling} formulated this problem as a trajectory optimization problem that maximizes an information metric under a budget constraint, and pointed out that such problems typically have high computational complexity (e.g., NP-hard / PSPACE-hard).
Traditional single-agent IPP methods can be broadly grouped into two categories: sampling-based methods and viewpoint/subgoal selection with continuous trajectory optimization. Representative sampling-based approaches, such as the RIG (Rapidly-exploring Information Gathering) family \cite{RIG/2014sampling,RIG2/2019sampling,RIG3/2018deterministic}, extend RRT/RRG-style planners to informative path planning by combining sampling search with branch-and-bound, enabling efficient trajectory search under continuous spaces and motion constraints while offering asymptotic optimality and scalability. In contrast, viewpoint/subgoal-based methods \cite{VP/2012/branch,vp2/2007nonmyopic,vp3/2020adaptive} typically first identify informative sensing targets and then generate feasible trajectories via local optimization or trajectory generation; for example, Hitz \textit{et al.} \cite{hitz/2017/adaptive} proposed a continuous-space IPP framework that combines Gaussian process modeling with evolutionary optimization and supports online replanning.

Recently, single-agent IPP has shifted from offline planning to adaptive IPP (AIPP). In this setting, deep reinforcement learning (DRL) is widely used to learn a mapping from belief states to actions, reducing online replanning cost. Recent studies further incorporate graph representations and attention mechanisms \cite{learning,catnipp} to improve global context modeling and mitigate the short-sightedness of local decision-making.
\subsection{Multi-Agent IPP}
Multi-agent informative path planning (MAIPP) remains less explored than its single-agent counterpart. Viseras \textit{et al.} \cite{viseras/2016/decentralized} addressed MAIPP by combining greedy planning with collision avoidance; however, prior studies in single-agent IPP have shown that greedy strategies often lead to short-sighted decisions and thus degrade long-term information-gathering efficiency, a limitation that is typically more pronounced in cooperative multi-agent settings \cite{hitz/2017/adaptive,RIG/2014sampling,MAIPPweakness/2017/randomized,popovic2020informative}. A promising direction is to extend effective single-agent IPP planners to the multi-agent setting and enable distributed coordination through Sequential Greedy Assignment (SGA) \cite{corah2017efficient}: agents plan their paths sequentially according to priority, and each subsequent agent explicitly conditions on the paths already assigned to higher-priority agents, yielding scalable cooperative planning. In recent years, deep reinforcement learning has also been introduced to MAIPP to reduce online computation by learning distributed coordination policies. Furthermore, intent-sharing and attention-based method \cite{Intent/MULTI/RL/YANG/2023,deolasee2024dypnipp} improve coordination by exchanging distributed predictions of future agent positions, but in adaptive settings with continuously updated beliefs, accumulated prediction errors may still degrade long-horizon planning performance.

\begin{figure*}[t]
    \centering
    \includegraphics[width=1 \textwidth, trim=20 10 10 5, clip]{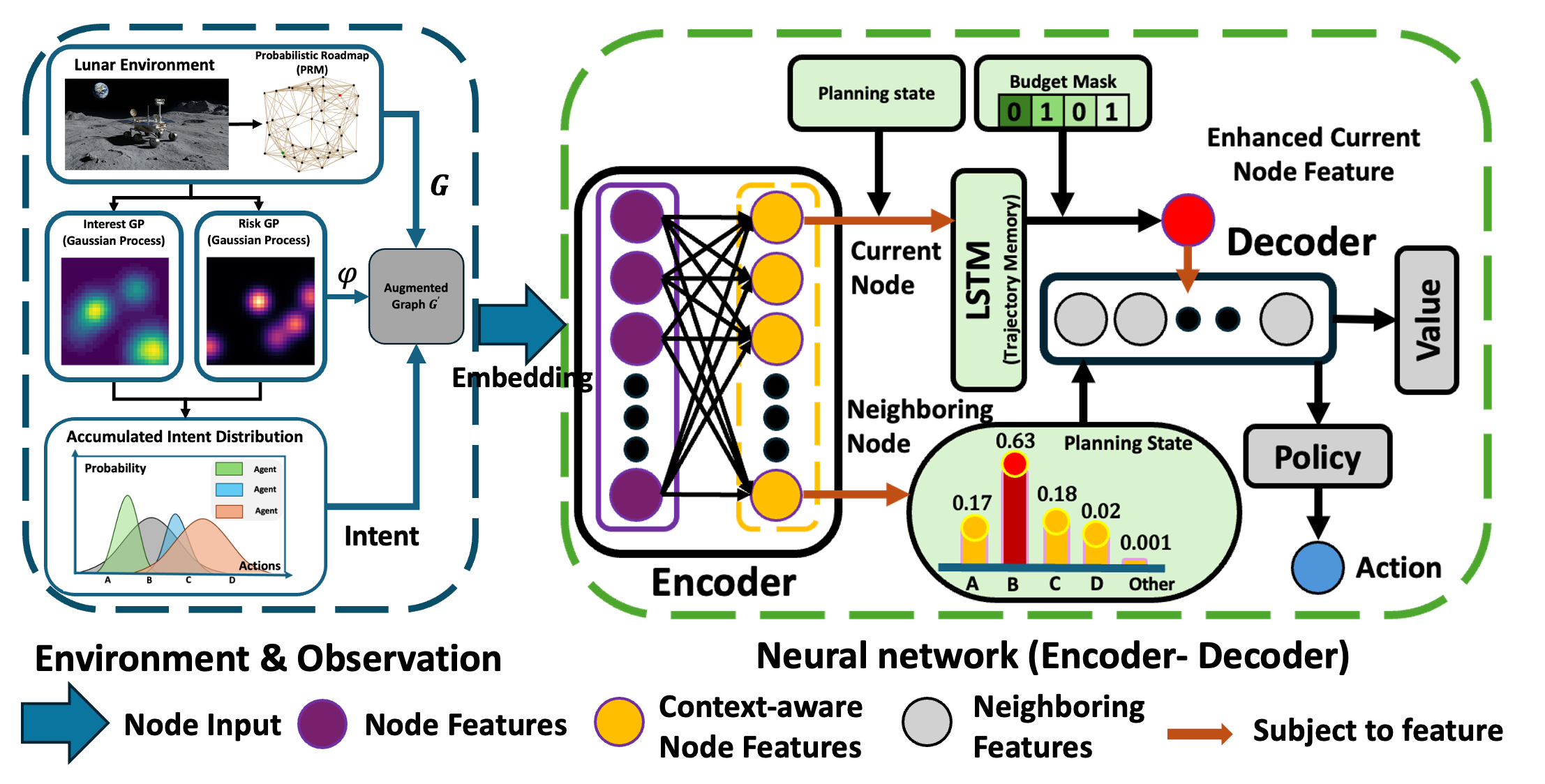}
    \caption{\textbf{Overview of our framework.} In the constructed map, our method gets two observation interest GP and risk GP, and mixes them with agents' intent to construct an augmented graph. Our neural network includes two main parts, the encoder and the decoder. After node inputting, the encoder relies on a self-attention block for noticing globe node belief and relationship, as context-aware node features. The decoder cares about the feature of the current node, neighboring node, planning state, and mask. Finally, input the value and action.     }
    \label{fig:framework}
\end{figure*}

\section{Background}
\subsection{Gaussian Process (GP)}
In informative path planning (IPP), both the latent high-value signal (interest) and terrain risk are modeled as continuous functions over a 2D workspace, $\zeta:\mathcal{E}\rightarrow\mathbb{R}$, where $\mathcal{E}\subset\mathbb{R}^2$ denotes the environment and is initially unknown to the agent(s). We employ a Gaussian process (GP) to approximate $\zeta$ by interpolating between discrete measurements, i.e., $\zeta \approx \mathcal{GP}(\mu, P)$, where $\mu$ and $P$ are the GP mean and covariance functions, respectively. Given $n$ measurement locations $X\subset\mathcal{E}$ with corresponding observations $Y$, and a set of query locations $X^{*}\subset\mathcal{E}$ at which predictions are required, the GP posterior mean $\mu$ and covariance $P$ are given by:
\begin{equation}
\mu=\mu(X^{*})+K(X^{*},X)\bigl[K(X,X)+\sigma_n^2 I\bigr]^{-1}\bigl(Y-\mu(X)\bigr)_,
\end{equation}
\begin{equation}
P=K(X^{*},X^{*})-K(X^{*},X)\bigl[K(X,X)+\sigma_{n}^{2}I\bigr]^{-1} \\
K(X^{*},X)^{T}_,
\label{eq:gp_cov}
\end{equation}
where $K(\cdot,\cdot)$ denotes a pre-selected kernel function (we use the Mat\'ern $3/2$ kernel), $\sigma_n^2$ is the measurement noise variance, and $I$ is the identity matrix.

\subsection{Multi-Agent Informative Path Planning (MAIPP)}
Given $m$ agents, each subject to an individual budget (path-length) constraint, the objective is to maximize the collective information gain. This problem can be formulated as:
\begin{equation}
\psi^{*}=\arg\max_{\psi\in\Psi}\sum_{i=1}^{m} I(\psi_i),
\quad \text{s.t.}\ C(\psi_i)\le B,\ 1\le i\le m,
\label{eq:maipp_obj}
\end{equation}
We seek an optimal set of trajectories $\psi^{*}=\{\psi_{1},\ldots,\psi_{m}\}$. Each trajectory $\psi_i$ yields an information gain $I(\psi_i)$ and incurs an execution cost $C(\psi_i)$, subject to shared team budget constraint $B$. In particular, we define $I(\psi_i)$ to reward uncertainty reduction over high-interest regions. These regions are identified using an upper confidence bound (UCB) criterion:
\begin{equation}
X_{I}=\left\{x_i \in X^{*}\ \middle|\ \mu_i^{-}+\beta P_{i,i}^{-}\ge \mu_{\mathrm{th}}\right\},
\label{eq:ucb_region}
\end{equation}
where $\mu_i^{-}$ and $P_{i,i}^{-}$ denote the GP prior mean and prior variance at location $x_i$, respectively, while $\mu_{\mathrm{th}}$ and $\beta\in\mathbb{R}^{+}$ control the selection threshold and confidence level. This formulation requires frequent measurements and online replanning to continuously reduce uncertainty within the high-interest region $X_I$.

\begin{figure*}[t]
    \centering
    \includegraphics[width=1 \textwidth, trim=0 10 10 5, clip]{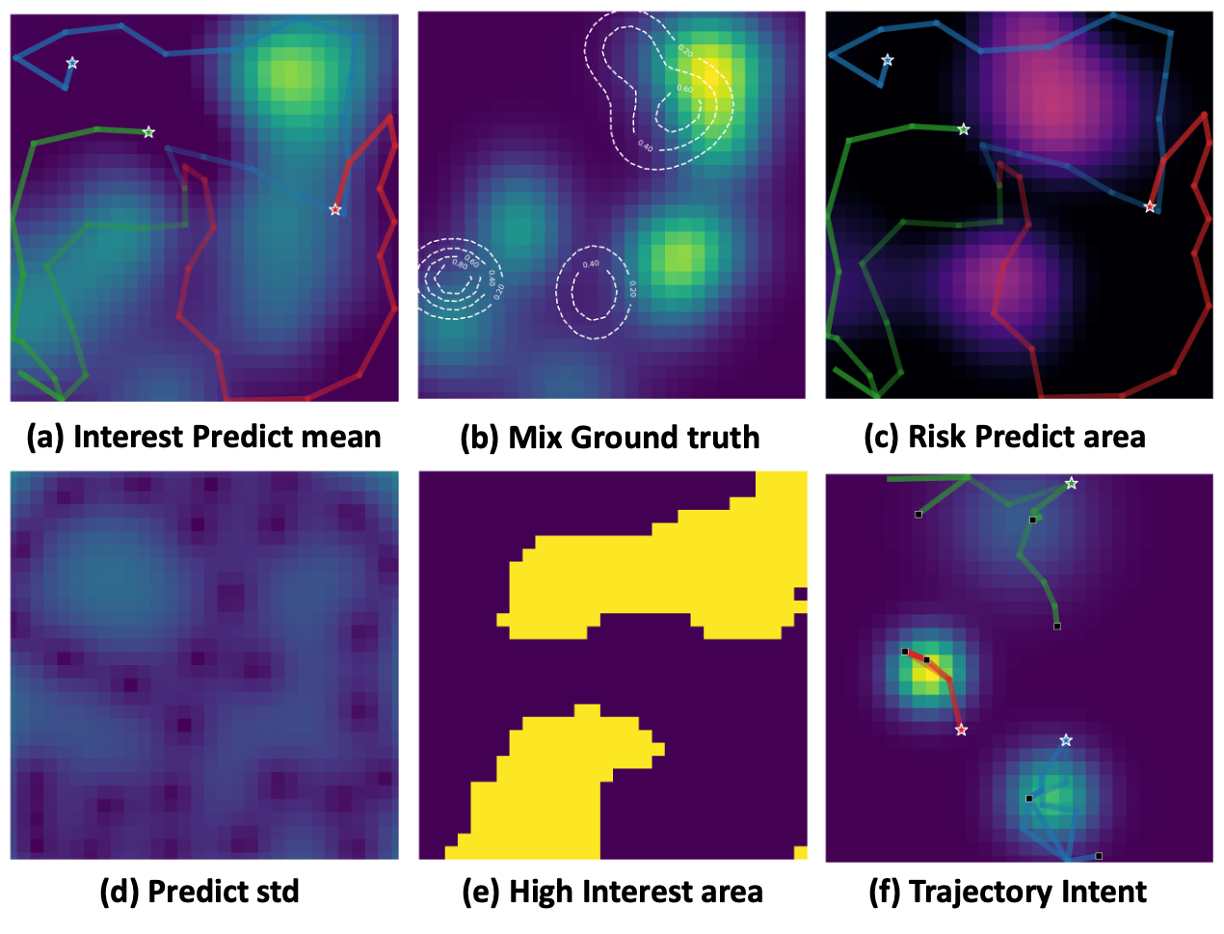}
    \caption{\textbf{Example of our method with 3 agents.} (a) shows the sampling data from three agents and predicts the global map, with trajectories in red, green, and blue. (b) represents the ground truth map of information distribution. (c) The opposite of (a) is the risk prediction map. (d) displays the GP predictive mean and standard deviation of information distribution. (e) highlights the regions of interest. (f) shows the agents' trajectory intent distribution.   }
    \label{fig:compare}
\end{figure*}

\section{Method}

\subsection{MAIPP as an RL Problem}

\textbf{Sequential Decision-Making Problem:}
We construct a roadmap graph \(G^i=(V^i,E^i)\) using a probabilistic roadmap (PRM) \cite{RIG2/2019sampling}. The vertex set \(V^i\) contains \(200\) nodes uniformly sampled over the workspace, and the edge set \(E^i\) is formed by connecting each node to its \(k\)-nearest neighbors. We assume an obstacle-free environment, so all sampled nodes are traversable, including nodes located in high-risk regions. Each node is represented by its 2D coordinates \(v^i=(x^i,y^i)\in V^i\).

All agents start from the same node and, following their policies \(\pi^i\), select neighboring nodes as next waypoints and move along straight-line edges at constant speed. Because decisions are executed in parallel, agents may reach waypoints at different times. During execution, each agent \(i\) builds a trajectory $
\psi^i=\{\psi^{i}_{1},\ldots,\psi^{i}_{n_i}\},
\psi^{i}_{j}\in\mathbb{R}^2,\ (\psi^{i}_{j},\psi^{i}_{j+1})\in E^i$, with path-length cost
$
C(\psi^i)=\sum_{j=1}^{n_i-1} L_2(\psi^{i}_{j},\psi^{i}_{j+1})$, where \(L_2(\cdot,\cdot)\) denotes Euclidean distance. Unlike per-agent budget formulations, we use a shared team budget:$ \sum_{i=1}^{m} C(\psi^i)\le B_{\text{team}}$.
Therefore, MAIPP is formulated as a sequential decision-making problem on the PRM over the graph set $G=\{G^1,G^2,\ldots,G^m\}$,
which supports flexible online replanning under a common budget pool.

\textbf{Dual-Agent Observation:}
To enable safe exploration, we maintain both an interest belief and a risk belief via Gaussian processes (see Fig.~\ref{fig:compare}(a),(c)). At decision step \(t\), the observation of agent \(i\) is
$s_t^{i}=\{G^{\prime i}_{it,t},\,G^{\prime i}_{rk,t},\,v_c^{i},\,B_c,\,\psi^{i},\,M_t^{i}\}$. The interest and risk beliefs are defined on the same roadmap topology, i.e., they share identical vertices and edges: $
G^{\prime i}_{it,t}=\bigl(V^{i},E^{i},\phi^{i}_{it,t}\bigr),
G^{\prime i}_{rk,t}=\bigl(V^{i},E^{i},\phi^{i}_{rk,t}\bigr)$.
Hence, both graphs have the same structure \((V^{i},E^{i})\) and differ only in node-wise attributes \(\phi(\cdot)\) induced by the corresponding GP beliefs. For node \(v_j^i\), the interest-graph node feature is
$\phi^{i}_{it,t}(v_j^i)=\bigl(\mu_{it,t}(v_j^i),\,\sigma_{it,t}(v_j^i),\,f_t^i(v_j^i)\bigr)$,
where \(f_t^i(v_j^i)\) denotes node-level intent features aggregated from other agents.
The risk graph is defined analogously using risk-belief features. In our implementation, risk is mainly injected through an upper-confidence estimate $\tilde{\mu}_{rk,t}(v_j^i)=\mu_{rk,t}(v_j^i)+\beta\,\sigma_{rk,t}(v_j^i)$,
which helps the policy explicitly trade off utility and safety. The executed trajectory of agent \(i\) is $\psi^{i}=(\psi_1^{i},\ldots,\psi_{c_i}^{i})$,
and under the shared-budget setting the remaining team budget is $B_c = B_{\text{team}}-\sum_{i=1}^{m} C(\psi^i)$.
Finally, \(M_t^{i}\) is a joint feasibility mask that filters actions violating constraints (e.g., budget infeasibility, node-collision constraints, and hard risk constraints).

\textbf{Action:} 
Agent $i$ follows a Dijkstra-based route \cite{dijkstra/2022/note} to arrive at a candidate node. Upon arrival, its belief, represented by a Gaussian process (GP), is updated by incorporating all measurements collected by the $m$ agents up to that time. Similarly, the intent is updated using the future predictions provided by the other $m-1$ agents. At each time step $t$, the updated information is fed into an attention-based neural network, which outputs a stochastic policy $\pi^i_\theta(\cdot \mid s_t^i)$ for selecting the next node to visit among neighboring candidates. The policy is parameterized by a set of weights $\theta$, and can be expressed as 
$
\pi^i_\theta\!\left(\psi^i_t = v^i,\ (v_s^i, v_d^i)\in E^i \,\middle|\, s_t^i\right)$, where $E^i$ denotes the edge set of the underlying graph.

\textbf{Reward:} To encourage efficient information gathering, at each decision step $t$ we define the reward as a linear combination of three components:
\begin{equation}
r_t \;=\; r_t^{\text{info}} \;-\; r_t^{\text{pen}} \;+\; r_t^{\text{term}} .
\end{equation}
The information-gain term $r_t^{\text{info}}$ measures the reduction in uncertainty induced by the current action. We use the trace of the GP posterior covariance as a scalar uncertainty measure (see Fig. \ref{fig:compare} (e)), $\Phi_t=\mathrm{Tr}(P_t)$, and define:
\begin{equation}
r_t^{\text{info}}=\max\!\left(0,\frac{\Phi_{t-1}-\Phi_t}{\Phi_{t-1}}\right).
\end{equation}
The penalty term $r_t^{\text{pen}}$ aggregates several event-based penalties: backtracking to recently visited nodes ($0.1$), multi-agent node collisions where multiple agents select the same next node ($0.2$), and budget overflow ($1.0$), which also terminates the episode. When the risk field is enabled (we set threshold is 0.7, means that not all risky areas are off-limits for exploration, as shown in Fig. \ref{fig:compare} (c)), an additional risk-related penalty is included (otherwise it is set to zero). Finally, $r_t^{\text{term}}$ is a terminal correction applied only at the end of the episode to better align the learning objective with the final IPP objective:
\begin{equation}
r_t^{\mathrm{term}} = -\lambda_{\mathrm{term}}\,\Phi_t\,\mathbb{I}\{s_{t+1}\in\mathcal{T}\},
\end{equation}
where $\Phi_t := \mathrm{Tr}(P_t)$ is the GP covariance trace, and
$\mathbb{I}\{s_{t+1}\in\mathcal{T}\}\in\{0,1\}$ is an indicator function that equals $1$ iff the transition at step $t$ enters the terminal set $\mathcal{T}$, and equals $0$ otherwise.
We set $\lambda_{\mathrm{term}} = 1/900$ to normalize the terminal penalty by the initial covariance trace under the $30\times 30$ discretization.

\subsection{Multi-Agents' Intent}
We represent each agent's predicted actions as a probabilistic distribution, referred to as the agent's \emph{intent} \cite{Intent/MULTI/RL/YANG/2023}. Specifically, the intent of agent $i$ at time step $t$ is modeled as a Gaussian distribution $\mathcal{GD}\!\left(\mu^{i}(t), \Sigma^{i}(t)\right)$, where $\mu^{i}(t)$ and $\Sigma^{i}(t)$ are the mean and covariance fitted from the planned trajectory of agent $i$ at time step $t$. Each agent continuously updates and broadcasts its intent, and fuses the intents of all other agents to support its own decision-making. Concretely, agent $i$ receives the intents from the other $m-1$ agents, aggregates them by summation, and normalizes the result to form an accumulated intent distribution. (see Fig. \ref{fig:compare} (f)) Agent $i$ then uses its own belief together with the accumulated intent to sample future paths and select actions, and subsequently broadcasts the updated $\mu^{i}(t)$ and $\Sigma^{i}(t)$. This low-communication design is well suited for real deployment in extreme lunar environments.

\subsection{Neural Network}
We build an attention-based neural network \cite{attention/2017} with an encoder--decoder architecture. Using attention layers as the fundamental building blocks, the network effectively captures inter-node dependencies in the augmented graph $G'$. The encoder learns context-aware node representations from $G'$ and provides the extracted features, together with the planning state and the budget mask \emph{M}, to the decoder. Conditioned on these inputs, the decoder outputs a policy $\pi$ that determines which neighboring node to visit next.

\textbf{Attention Layer:}
The layer takes a query source $h^{q}$ and a key--value source $h^{k,v}$ as inputs, and updates each query feature by aggregating value vectors with similarity-based weights.
Specifically, linear projections are first applied to obtain queries, keys, and values:
\begin{equation}
\begin{aligned}
q_i &= W^{Q} h_i^{q},\quad
k_i = W^{K} h_i^{k,v},\quad
v_i = W^{V} h_i^{k,v}, \\
u_{ij} &= \frac{q_i^{T} \cdot k_j}{\sqrt{d}},\quad
w_{ij} = \frac{e^{u_{ij}}}{\sum_{j=1}^{n} e^{u_{ij}}}, h'_i = \sum_{j=1}^{n} w_{ij} v_j ,
\end{aligned}
\end{equation}
where $W^{Q}$, $W^{K}$ , and $W^{V}\subset\mathbb{R}^{d\times d}$ are learnable parameter matrices and $d$ denotes the feature dimension.

\textbf{Encoder:} The encoder captures inter-node dependencies in the augmented graph $G'$. 
Each node in $V'$ is projected to a $d$-dimensional embedding and augmented with Laplacian positional encodings. 
We then apply self-attention ($h^{q}=h^{k,v}=h^{n}$) to obtain context-aware node features $h^{en}$ for downstream decoding.

\textbf{Decoder:} The decoder takes the current-node embedding and its neighbor embeddings from $h^{en}$ (all in 128 dimensions). 
It concatenates the interest threshold and remaining budget to the current embedding, then projects back to 128 dimensions.
An LSTM \cite{LSTM/1997/long} encodes trajectory history to form an enhanced current representation. 
A pointer-style attention layer \cite{pointer/2015} finally outputs normalized attention weights, defining the stochastic policy $\pi_\theta(\cdot \mid s_t)$ over neighboring nodes.

\subsection{Training}
We train the policy using Proximal Policy Optimization (PPO) \cite{PPO/2017/proximal}. At the beginning of each training episode, we randomly sample and average \(8\!-\!12\) two-dimensional Gaussian components over \([0,1]^2\) to construct the interest ground truth \(y_{it}\). In addition, we sample \(4\!-\!6\) Gaussian components to generate the risk ground truth \(y_{rk}\). We combine them as $y_{mix}=y_{it}\cdot\bigl(1-\lambda\,y_{rk}\bigr)$,
with \(\lambda=0.5\) (see Fig.~\ref{fig:compare}(b)), which helps the agents capture the interaction between utility and risk. Both the ground-truth map and the belief map are discretized on a \(30\times 30\) grid. Therefore, the initial covariance trace (empty prior belief) is \(900\). Agent start positions are sampled uniformly from \([0,1]^2\). For each episode, a PRM with \(200\) nodes is constructed, and each node is connected to its \(k=20\) nearest neighbors. We train with \(m=3\) agents under a shared team budget \(B_{\text{team}}\) (instead of per-agent independent budgets). Each agent acquires a new measurement after traveling a distance of \(0.2\) since its previous measurement, and an episode terminates when the shared budget is exhausted (equivalently, when no agent can continue under feasibility constraints). We use Adam with learning rate \(1\times 10^{-4}\), decayed by a factor of \(0.96\) every \(32\) steps, and a batch size of \(256\). PPO performs \(8\) update epochs per training step. Training is conducted on a workstation with a 16-core AMD EPYC 7542 CPU and four NVIDIA GeForce RTX 3090 GPUs, and converges in approximately \(24\) hours (\(\sim 17{,}000\) episodes).

\section{Comparison and Analysis}

In all experiments, the agents share the same initial position $v_1$ and the same budget $B$. Based on this setting, we aggregate each agent's sampled predicted trajectories $\{g^i(t)\}$ into \textbf{trajectory intent} (TI), where the intent distribution is obtained by fitting a Gaussian distribution (GD) over all nodes along the sampled trajectories. A visual example of the resulting intent representation is shown in Fig.~\ref{fig:compare}.

\begin{table*}[t]
\caption{Comparison between different domain-aware variants in terms of $\mathrm{Tr}(P_f)$ for 3 agents (10 trials on 30 instances).}
\label{tab:1}
\centering
\resizebox{\textwidth}{!}{%
\begin{tabular}{l|cc|cc|cc|cc|cc}
\hline
\multirow{2}{*}{Method} 
& \multicolumn{2}{c|}{Domain} 
& \multicolumn{2}{c|}{Budget 2} 
& \multicolumn{2}{c|}{Budget 3} 
& \multicolumn{2}{c|}{Budget 4} 
& \multicolumn{2}{c}{Budget 5} \\
& Interest & Risk & mean & std & mean & std & mean & std & mean & std \\
\hline
SGA + RRT (0.9,1.0) & $\checkmark$     & $\times$      & 182.53 & 103.83 & 139.49 & 96.95 & 97.45 & 86.52 & 44.64 & 35.89 \\
SGA + RRT (0.3,0.4) & $\checkmark$      & $\times$      & 200.46 & 203.52 & 270.32  & 302.05 & 154.26  & 252.81 & 214.67  & 219.22 \\
CAtNIPP + Greedy    & $\checkmark$      & $\times$      & 45.59  & 35.38 & 26.27  & 31.21 & 21.48   & 31.56 & 23.42  & 31.13 \\
CAtNIPP + TI (8,5)  & $\checkmark$  & $\times$      & 48.57  & 33.87 & 25.51   & 21.34 & 26.18   & 18.32 & 14.24   & 7.08 \\
\textbf{Ours} + TI (8,5) & $\checkmark$  & $\checkmark$  & \textbf{37.30} & 16.26 & \textbf{21.76} & 12.34 & \textbf{19.32} & 10.24 & \textbf{10.99} & 6.16 \\
\hline
\end{tabular}%
}
\end{table*}

\begin{table}[t]
\caption{Ablation study in terms of final average $\mathrm{Tr}(P_f)$ for 10 agents (10 trials on 30 instances, single-domain).}
\label{tab:performance_10_agents}
\centering
\resizebox{\columnwidth}{!}{%
\begin{tabular}{l|cc|cc|cc}
\hline
\multirow{2}{*}{Method} 
& \multicolumn{2}{c|}{Budget 2} 
& \multicolumn{2}{c|}{Budget 3} 
& \multicolumn{2}{c}{Budget 4} \\
& mean & std & mean & std & mean & std \\
\hline
SGA + RRT (0.9,1.0)   & 32.63 & 1.82 & 21.23 & 1.78 & 15.26 & 1.14 \\
CAtNIPP + Greedy      & 29.56 & 3.80 & 12.02 & 1.22 & 6.35  & 0.83 \\
CAtNIPP + TI (8,5)    & \textbf{18.03} & 2.26 & 5.32 & 0.46 & \textbf{3.30} & 0.34 \\
\textbf{Ours} + TI (8,5) & 19.69 & 2.35 & \textbf{5.21} & 0.23 & 3.07 & 0.34 \\
\hline
\end{tabular}%
}
\end{table}

\subsection{Experimental Setup}

We compare our method with two sequential greedy assignment (SGA) baselines, namely \textbf{SGA-RRT} \cite{firstIPP/2013/sampling} and \textbf{Greedy-CAtNIPP} \cite{catnipp}, as well as the \textbf{intent-based version of CAtNIPP} \cite{Intent/MULTI/RL/YANG/2023}. 
(i) \textbf{SGA-RRT} \cite{firstIPP/2013/sampling} uses a conventional RRT-based planning strategy \cite{RRT,RRT2}, where agents plan sequentially by conditioning on both the current belief and the paths assigned to higher-priority agents. Each agent samples a set of candidate paths and selects the one that minimizes $\mathrm{Tr}(P_f)$. In execution, only a short segment of the selected path is followed (set to 0.2 in our experiments), after which the belief is updated and replanning is performed. 
(ii) \textbf{Greedy-CAtNIPP} \cite{catnipp} selects the nearest viewpoint at each decision step. 
(iii) \textbf{Intent-CAtNIPP} \cite{Intent/MULTI/RL/YANG/2023} extends \textbf{CAtNIPP} \cite{catnipp} to the multi-agent setting by incorporating intent information.

We report the final uncertainty $\mathrm{Tr}(P_f)$ in the experiment. $\mathrm{Tr}(P_f)$ is defined as the trace of the GP posterior covariance on the query grid and equals the sum of predictive variances over all grid locations. Lower values indicate more effective uncertainty reduction. The reported \textit{mean} and \textit{std} are computed over repeated runs. The \textit{mean} summarizes the expected terminal uncertainty and the \textit{std} reflects the dispersion across runs and thus the stability of performance under the same evaluation protocol.

In the Table~\ref{tab:1}, we set the number of agents to $m=3$, and all methods are evaluated under the same budget settings. To mimic hazardous lunar terrain, we randomly generate 4--6 risk zones in each instance. Since the baselines do not model risk explicitly, risky waypoints are removed from the sampled roadmap nodes for those methods. In contrast, our method explicitly models both interest and risk, and we additionally report a risk-free variant in Table~\ref{tab:performance_10_agents}. 

For the RRT baselines, RRT$(a,b)$ denotes that the selected trajectory length lies in the range $[a,b]$. For the TI-based variants, $(a,j)$ denotes that each agent samples $a$ candidate trajectories and each trajectory contains $j$ nodes. To further evaluate robustness under communication constraints, we also test communication ranges of 0.3 and 0.6, and compare them with the global communication setting in Table~\ref{tab:limited_communication}.

\subsection{Result and Analysis}

Our method performs best in the risk-aware three-agent setting and remains competitive across the other evaluated settings. Compared with the sampling-based \textbf{SGA-RRT} \cite{firstIPP/2013/sampling} and the local greedy baseline \textbf{Greedy-CAtNIPP} \cite{catnipp} , our method achieves substantially lower final uncertainty, showing the benefit of jointly modeling trajectory intent and environmental risk.

In the Table~\ref{tab:1}, all methods generally benefit from larger budgets, while our method consistently achieves the best performance. For example, our method reduces $\mathrm{Tr}(P_f)$ from 37.30 to 10.99 as the budget increases from 2 to 5, whereas \textbf{Greedy-CAtNIPP} \cite{catnipp} only decreases from 45.59 to 23.42 over the same range. Compared with \textbf{Intent-CAtNIPP} \cite{Intent/MULTI/RL/YANG/2023}, the gain of our method is moderate but consistent; for instance, the result improves from 25.51 to 21.76 at budget 3 and from 14.24 to 10.99 at budget 5. The gap becomes much larger when compared with \textbf{SGA-RRT} \cite{firstIPP/2013/sampling}, whose results remain above 100 in several settings. These results suggest that trajectory-intent modeling already improves coordination, while explicitly incorporating risk information further enhances planning quality in hazardous environments.

\begin{table}[t]
\caption{Performance under limited communication ranges, in terms of $\mathrm{Tr}(P_f)$, for 3 agents (10 trials on 30 instances).}
\label{tab:limited_communication}
\centering
\resizebox{\columnwidth}{!}{%
\begin{tabular}{l|c|cc|cc|cc}
\hline
\multirow{2}{*}{Method} & \multirow{2}{*}{\shortstack{Communication\\Range}}
& \multicolumn{2}{c|}{Budget 2}
& \multicolumn{2}{c|}{Budget 3}
& \multicolumn{2}{c}{Budget 4} \\
& & mean & std & mean & std & mean & std \\
\hline
SGA + RRT (0.3,0.4)  & \multirow{4}{*}{0.3}  & 110.32 & 63.57 & 64.34 & 61.99 & 39.74  & 32.59 \\
CAtNIPP + Greedy     &                        & \textbf{35.24} & 12.34 & \textbf{18.04} & 11.82 & \textbf{16.21} & 8.27 \\
CAtNIPP + TI (8,5)   &                        & 60.99 & 23.07 & 21.39 & 18.65 & 25.12 & 18.65 \\
\textbf{Ours} + TI (8,5) &                    & 40.91 & 21.07 & 22.18 & 15.65 & 20.12 & 18.42 \\
\hline
SGA + RRT (0.3,0.4)  & \multirow{4}{*}{0.6}  & 154.02 & 72.61 & 152.36 & 62.54 & 148.87 & 61.05 \\
CAtNIPP + Greedy     &                        & \textbf{45.02} & 32.91 & 40.94 & 12.54 & 38.87 & 10.75 \\
CAtNIPP + TI (8,5)   &                        & 51.70 & 3.72 & 38.80 & 29.73 & 27.61 & 21.07 \\
\textbf{Ours} + TI (8,5) &                    & 46.67 & 23.07 & \textbf{24.39} & 14.65 & \textbf{22.12} & 14.92 \\
\hline
\textbf{Ours} + TI (8,5) & Global             & 37.30 & 16.26 & 21.76 & 12.34 & 19.32 & 10.24 \\
\hline
\end{tabular}%
}
\end{table}

Table~\ref{tab:performance_10_agents} shows that in the single-domain setting without risk, our method and \textbf{Intent-CAtNIPP} \cite{Intent/MULTI/RL/YANG/2023} achieve very similar results and both clearly outperform the conventional baselines. For example, at budget 3 the two methods obtain 5.21 and 5.32, respectively, and at budget 4 they further improve to 3.07 and 3.30. By contrast, \textbf{SGA-RRT} \cite{firstIPP/2013/sampling} still yields 21.23 and 15.26 under the same budgets. This indicates that trajectory-intent modeling is the main source of performance gain in the risk-free setting, while the additional benefit of our method becomes more evident when explicit risk constraints are introduced.

Table~\ref{tab:limited_communication} further shows that reducing the communication range degrades the performance of all methods, especially for \textbf{SGA-RRT} \cite{firstIPP/2013/sampling}. At communication range 0.3, \textbf{Greedy-CAtNIPP} \cite{catnipp}  performs best, while our method remains competitive and still substantially outperforms \textbf{SGA-RRT} \cite{firstIPP/2013/sampling} (e.g., 22.18 versus 64.34 at budget 3). When the range increases to 0.6, our method becomes the best-performing method at higher budgets, achieving 24.39 and 22.12 at budgets 3 and 4, respectively, compared with 40.94 and 38.87 for \textbf{Greedy-CAtNIPP} \cite{catnipp} . Overall, these results indicate that our method is robust to communication degradation, although global communication still provides the best overall performance.

\begin{figure}[t]
\centering
\includegraphics[width=1\linewidth, trim=0 0 10 0, clip]{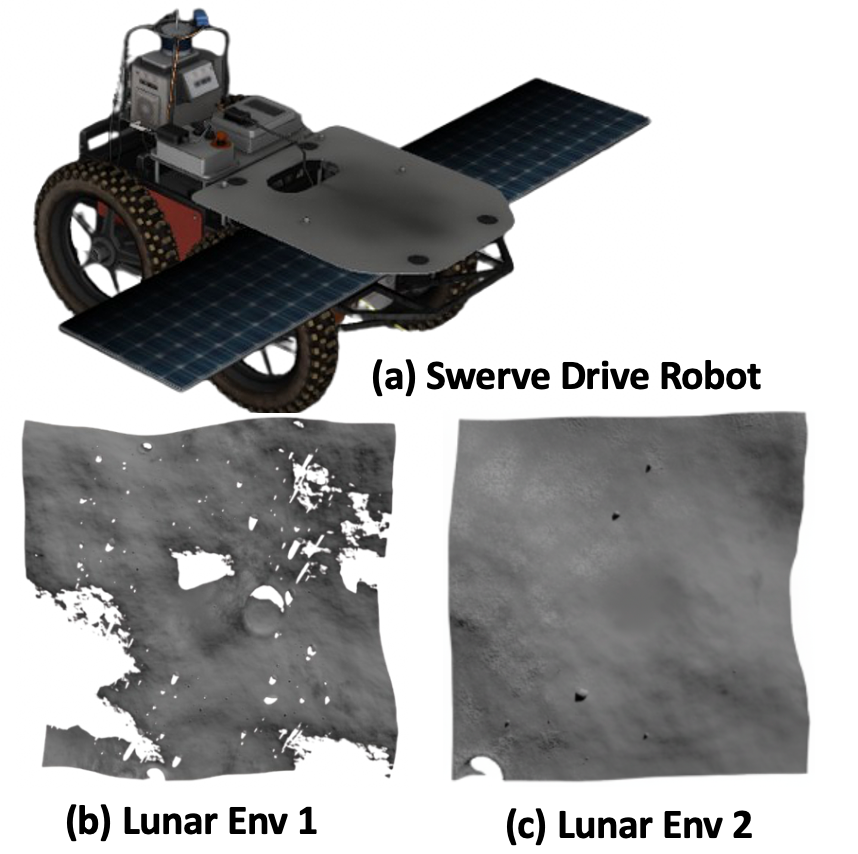}
\vspace{-15pt}
\caption{\textbf{Simulation and validation.} Two lunar environments with different terrain.}
\label{fig:simulation}
\vspace{-15pt}
\end{figure}

\section{Simulation and Validation}
We constructed a lunar multi-robot simulation scenario in Gazebo to evaluate the adaptability and effectiveness of the proposed framework in complex unstructured terrains. The simulation system consists of three identical swerve-drive rovers (Fig.~\ref{fig:simulation}(a)) operating collaboratively under a shared team budget. We further design two lunar environments with distinct geomorphology characteristics (Fig.~\ref{fig:simulation}(b,c)), referred to as Lunar Environment~1 and Lunar Environment~2. Both environments include variations in surface appearance, terrain undulations, obstacle layouts, and local irregularities to better approximate the uncertainty of off-world operations. Lunar Environment~1 exhibits stronger terrain fluctuations and more uneven hazard patterns, making it suitable for evaluating risk-aware exploration behavior (see Fig.~\ref{fig:motivation}). In contrast, Lunar Environment~2 has a smoother global structure while retaining noticeable local perturbations, which is useful for assessing exploration robustness under weak visual cues and limited observability.

To focus on informative exploration and multi-agent coordination, we adopt a 2D exploration abstraction: planning and belief updates are performed on a planar workspace, while the influence of 3D terrain (e.g., slope-induced slip) is summarized by a scalar risk field defined over the same 2D domain. Both the interest field (for sparse evidence) and the risk field are treated as latent continuous functions and are estimated online using Gaussian-process beliefs (interest GP and risk GP). During execution, each rover acquires only local measurements at visited waypoints, and the team fuses the accumulated measurements through communication to update the shared GP beliefs. The underlying latent fields are used only for simulation-based observation generation and offline evaluation, and are never exposed to the policy as privileged information. Comparative experiments in these two environments therefore assess cooperative exploration efficiency, risk-aware decision making, and robustness under different budgets and communication ranges.

\section{CONCLUSIONS}

In summary, multi-agent exploration in lunar environments faces major challenges, including communication limitations, complex terrain, hazardous regions, and sparse targets that are difficult to observe. These factors make it difficult for conventional methods based on local greedy strategies or single-source information to achieve high search efficiency, effective coordination, and operational safety at the same time. To address these challenges, this article proposes a deep reinforcement learning framework for multi-agent informative path planning (MAIPP). By incorporating a self-attention mechanism, the proposed framework improves the modeling of global environmental information and inter-agent interactions, thereby enhancing sequential decision-making in complex environments. Our approach simultaneously maintains and updates two beliefs, namely interest and risk, allowing agents to balance information gain and exploration safety. In addition, we incorporate a trajectory belief during training to better model teammates' future motions and reduce redundant exploration. Experimental results demonstrate that the proposed method achieves strong performance in information gathering, risk mitigation, and cooperative exploration. Although the current study is mainly validated in simulation, this work provides a feasible step toward more autonomous and reliable multi-agent lunar exploration. As Armstrong once said, many small advances can eventually lead to a giant leap in technology.
\addtolength{\textheight}{-12cm}   





\bibliographystyle{IEEEtran}
\bibliography{mybib}

\end{document}